\documentclass[agriengineering,article,accept,moreauthors]{Definitions/mdpi} 
\firstpage{1} 
\pubvolume{8}
\issuenum{10}
\articlenumber{410}
\pubyear{2026}
\copyrightyear{2026}
\externaleditor{Dongqiao Peng} % More than 1 editor, please add `` and '' before the last editor name
\datereceived{14 July 2026} 
\daterevised{18 September 2026} % Comment out if no revised date
\dateaccepted{20 September 2026} 
\datepublished{30 September 2026} 
\usepackage{makecell}
\newcommand{\re}[1]{\textcolor{black}{#1}}

\Title{Cattle-CLIP: A Multimodal Framework for {Dairy} %%\textcolor{( The title is different from the one submitted online at susy.mdpi.com. Please confirm which one is correct.)}
 Cattle Behaviour Recognition from Video}

\Author{{Huimin Liu $^{1}$\orcidA{},} %\textcolor{( 1. Please carefully check the accuracy of names and affiliations.  2. Orcid number is not mentioned on susy, please confirm if keep it in tex.)}
 Jing Gao $^{2}$, Daria Baran $^{1}$\orcidB{}, Axel X{.} %\textcolor{(  We added dots to abbreviate the middle names. Please confirm and please add dots to the abbreviated middle name on SuSy to keep this consistent. Same as below.)}
 Montout $^{1}$, Neill W{.} Campbell $^{2}$\orcidD{} and Andrew W{.} Dowsey $^{1,}$*\orcidC{}}

\AuthorNames{Huimin Liu, Jing Gao, Daria Baran, Axel X. Montout, Neill W. Campbell and Andrew W. Dowsey}

\address{%
$^{1}$ \quad Bristol Veterinary School, University of Bristol, {Bristol BS40 5DU, UK;} %\textcolor{( 1. We revised the country format, please confirm. Same as in aff. 2. 2. Please add city and postcode.  Same as in aff. 2. )}
sf24225@bristol.ac.uk (H.L.);\linebreak daria.baran@bristol.ac.uk (D.B.); axel.montout@bristol.ac.uk (A.X.M.) \\
$^{2}$ \quad School of Computer Science, University of Bristol, {Bristol BS8 1UB, UK;} 
jing.gao@bristol.ac.uk (J.G.); neill.campbell@bristol.ac.uk (N.W.C.)}

\corres{Correspondence: andrew.dowsey@bristol.ac.uk; Tel.: {+44-11-7455-0890} %\textcolor{( We revised the Tel. format, please confirm. )}
}

\abstract{Monitoring cattle behaviour provides valuable insights into animal health, productivity and welfare conditions. 
However, robust behaviour recognition in real-world farm environments remains challenging due to several data-related limitations, including the scarcity of well-annotated livestock video datasets and the substantial domain gap between large-scale pretraining corpora and agricultural surveillance footage. 
\re{To address these challenges, Cattle-CLIP, a domain-adaptive vision--language framework based on Contrastive Language-Image Pretraining (CLIP), is proposed, in which cattle behaviour recognition is reformulated as a cross-modal semantic alignment task rather than a purely visual classification problem.}
Instead of directly fine-tuning visual backbones, Cattle-CLIP incorporates a temporal integration module to extend image-level contrastive pretraining to video-based behaviour understanding, enabling consistent semantic alignment across time.
In order to minimise the domain discrepancy between web-scale pretraining data and real-world cattle surveillance scenarios, customised data augmentation and specialised behaviour-oriented prompts are employed.
\re{Furthermore, {CattleBehaviours6,} %\textcolor{(Please confirm if the italic of paragraph is unnecessary and can be removed, please check and confirm all italics in whole tex.)}
 a curated and behaviour-consistent video dataset comprising 1905 annotated clips across six indoor behaviours was constructed: \textit{feeding}, \textit{drinking}, \textit{standing-self-grooming}, \textit{standing-ruminating}, \textit{lying-self-grooming}, \textit{lying-ruminating}, to support model training and evaluation.}
Beyond serving as a benchmark for our proposed method, the dataset provides a standardised ethogram definition, offering a practical resource for future research in livestock behaviour analysis.
The performance of Cattle-CLIP was assessed under both supervised and few-shot learning paradigms, particularly targeting behaviour recognition in data-constrained environments, an important yet insufficiently studied task in livestock surveillance. 
Experimental evaluation demonstrated that the proposed model reached 96.1\% accuracy across six behaviours in supervised tasks, while achieving both high precision and recall for \textit{feeding}, \textit{drinking} and \textit{standing-ruminating} behaviours.
Moreover, the model maintained promising few-shot performance under limited labelled data, highlighting the potential of multimodal learning \re{for data-constrained cow behaviour recognition}.}


\begin{document}

%%%%%%%%%%%%%%%%%%%%%%%%%%%%%%%%%%%%%%%%%%
\section{Introduction}

Cattle play a vital role in global agriculture, particularly as the primary source of milk production.
The global market for dairy foods was valued at \$947.11 billion in 2024 \citep{fortune_dairy_foods_market_2025}. 
As a result, cattle farming is practised worldwide across diverse production systems. 
Cattle behaviour has been increasingly recognised as a valuable indicator of health status, productivity and overall welfare \citep{borchers2017machine, rushen2012automated}, drawing substantial attention from farmers, veterinarians and researchers alike. Behavioural changes in cattle, such as exploratory movements, reproductive activities, feeding and drinking habits and other social interactions, are commonly regarded as early indicators of health disorders (e.g., mastitis) as well as physiological states such as oestrus and pregnancy \citep{qiao2022c3d, caplen2021changes}. Therefore, monitoring and recognising cattle behaviour is essential for informing timely management decisions to optimise production efficiency, enhance welfare and hence improve environmental sustainability in precision livestock farming.

Substantial research on the recognition of cattle behaviours with precision livestock farming technologies has been carried out in recent years. Current approaches are commonly categorised into three types based on their data acquisition sources: contact-based, non-contact and hybrid methods.
Among these, contact-based systems utilise wearable sensors, including inertial measurement units (IMUs), to gather behavioural data for subsequent processing.
Shen et al. \citep{shen2020automatic} recognised the ingestive-based behaviour of cows using a triaxial accelerometer sensor to collect jaw-movement data. 
Similarly, Tian et al. \citep{tian2021real} used geomagnetic and acceleration sensors to collect cow behavioural data such as feeding, ruminating, drinking, resting, running, walking and head-shaking.  
A fully convolutional network was utilised by Liu et al. \citep{liu2023classification} for automatic recognition of seven cow behaviours, namely feeding, rub scratching (leg and neck), lying-ruminating, lying, self-licking and social licking, based on IMU-based data. 
Although the contact-based approaches can provide high precision, they can induce stress in animals and are susceptible to external factors such as abrasion, moisture infiltration and hardware degradation, which affect the longevity and reliability of the sensing devices \citep{wu2021using}.

Hybrid approaches have also been explored to combine the complementary information provided by wearable sensors and vision-based systems. For instance, the multimodal data fusion algorithm proposed by \citep{zhang2024new} combines information from neck-mounted IMUs and video data captured by barn-installed cameras. This approach achieved a high recognition accuracy of 98.80\% across five behaviours: drinking, feeding, lying, standing \mbox{and walking.}

To avoid the need for wearable devices altogether, fully non-contact methods utilising cameras to collect data have emerged as stress-free \citep{hlimi2024application} and cost-effective technologies, which have been widely applied for animal behaviour recognition \citep{mcdonagh2021detecting}. 
Guo et al. \citep{guo2019detection} integrated spatial and motion features via inter-frame differencing and background subtraction to recognise cow mounting behaviour. A support vector machine classifier was adopted for final recognition. Despite achieving 98.3\% accuracy on a dataset of 30 videos, the method’s dependence on handcrafted features and traditional classifiers would restrict adaptability to diverse scenes or large-scale datasets.
In \citep{ma2022basic}, an enhanced {R(2+1)D-based} %\textcolor{(Please confirm if needs to add spaces at two side of +, if is, please revise in whole tex.)}
 network was developed for video-based recognition of fundamental motion behaviours such as lying, standing and walking, yielding 95\% overall accuracy. Nevertheless, both methods are limited in scope as they focus on either a single behaviour or coarse-grained motion classes and lack the capacity to distinguish more nuanced or ambiguous behaviours.

Driven by the fast development of computer hardware and computational resources, deep learning models have gradually supplanted traditional machine learning approaches for behaviour recognition tasks. Among these, CNN-based and LSTM-based models have shown strong performance by capturing both spatial structure and temporal dynamics from video data \citep{feichtenhofer2019slowfast, arif2021bidirectional, wu2023can}. 
Originally introduced for natural language processing tasks, Transformer-based approaches have recently shown strong capability across diverse computer vision domains, including classification, detection and segmentation \citep{carion2020end, perrett2021temporal, piergiovanni2023rethinking, strudel2021segmenter}. Their strength lies in their ability to model long-range dependencies, making them particularly well suited for capturing temporal dynamics. Transformer-based models have been successfully applied to various animal behaviour recognition tasks, such as recognising aggressive and non-aggressive behaviours in pigs \citep{souza2024pig}, identifying scent marking, exploring, walking, resting and sniffing behaviours in wild giant pandas \citep{liu2024joint} and classifying cattle individual and social behaviours \citep{li2024dynamic}.
These applications underscore the expanding role of Transformer architectures in improving animal behaviour analysis methods.

The introduction of Transformer-based architectures has played a pivotal role in the rapid evolution of vision--language models (VLMs) by introducing a unified framework for processing heterogeneous data types. VLMs, such as Contrastive Language-Image Pretraining (CLIP), ALIGN and Flamingo \citep{radford2021learning, jia2021scaling, alayrac2022flamingo},
leverage the semantic richness of language and the increasing availability of extensive multimodal datasets to achieve strong performance across a variety of learning paradigms, including fully supervised, \mbox{few-shot \citep{tian2020rethink}} and zero-shot \citep{8237711} scenarios.
Among them, CLIP has become a widely adopted classification framework that reformulates classification as a matching problem between visual features and textual descriptions, rather than encoding category labels as arbitrary numerical identifiers or one-hot vectors.
Since large-scale video--language datasets remain scarce due to the resource-intensive nature of their collection, many studies have instead focused on bridging pretrained image--language models with video representation learning and demonstrated their generalisation capabilities in video recognition tasks on human action datasets \citep{wang2021actionclip, ni2022expanding, rasheed2023fine} and wildlife behaviour datasets \citep{jing2023category, gabeff2024wildclip}. 
Although these approaches have also achieved significant performance improvements on Kinetics-400 \citep{kay2017kinetics} compared with both 3D CNN-based methods and unimodal Transformer-based approaches, their application to cattle behaviour analysis remains largely unexplored.
Existing 3D-CNN-based methods typically rely on large-scale annotated video datasets to learn effective spatio-temporal representations, limiting their generalisation capability in agricultural settings where labelled livestock video data are scarce and expensive to obtain.
In contrast, multimodal pretraining aligns naturally with livestock monitoring tasks, as semantic supervision from image-text representations can improve sample efficiency under limited-data conditions while providing a flexible basis for extending recognition to new or under-represented~behaviours.

{Cattle-CLIP is proposed as a multimodal framework that integrates textual semantics with video representations for dairy cow behaviour recognition. 
To extend the image encoder of CLIP for modelling temporally consistent behavioural patterns, a lightweight temporal integration module aggregates frame-level features across time, enabling the representation of subtle and persistent behavioural dynamics.
In addition, customised data augmentation strategies are introduced to improve robustness to variations in animal posture, spatial deformation and camera viewpoint commonly encountered in farm surveillance videos. Domain-specific behaviour prompts are further incorporated to preserve behaviour-relevant semantic information and strengthen correspondence between the visual and language modalities.
Beyond fully supervised recognition, Cattle-CLIP also addresses data-scarce scenarios in which both well-represented base behaviours and under-represented novel behaviours must be recognised simultaneously. A base-to-novel transfer strategy is therefore introduced to transfer semantic knowledge from well-characterised base behaviours to novel behaviours, while a replay mechanism \citep{rebuffi2017icarl} is used to mitigate catastrophic forgetting \citep{aleixo2023catastrophic} and maintain recognition of previously learned behaviours.}

\re{Accordingly, this study has two main objectives. First, it aims to develop and evaluate Cattle-CLIP, a multimodal vision--language framework for recognising six indoor dairy-cow behaviours from surveillance video. Specifically, this study investigates whether temporal feature integration, domain-specific visual augmentation and behaviour-oriented textual prompts improve supervised behaviour recognition, and whether the proposed base-to-novel transfer strategy supports recognition of novel behaviours under limited labelled data while maintaining performance on previously learned behaviours. Second, this study aims to construct and publicly release CattleBehaviours6, a curated video dataset with standardised behaviour definitions and benchmark data partitions, providing a reusable resource for the development and comparison of cow behaviour recognition methods. Detailed information on behaviour definitions, data collection and annotation is provided in Section \ref{dataset curation}.}

% \textbf{Dataset}
% Public datasets are fundamental for advancing deep learning research, however annotating cattle behaviour requires domain expertise and sustained observation, making large-scale dataset construction resource-intensive. 
% Consequently, publicly available cattle behaviour datasets remain limited, with the exception of the CVB \citep{zia2023cvbvideodatasetcattle} and CBVD-5 \citep{li2024new}. 
% The CVB dataset captures behavioural videos in outdoor environments using unmanned aerial vehicles, while the CBVD-5 dataset records five cattle behaviours from 107 barn-housed cows. However, they both are annotated in the AVA format \citep{gu2018ava}, primarily designed for spatio-temporal detection tasks rather than classification. 
% To facilitate advancements in the domain of cattle research, we release CattleBehaviours6, which consists of six basic individual behaviours: \textit{feeding, drinking, self-grooming (standing and lying) and ruminating (standing and lying)}, together with a standardised ethogram definition for consistent behaviour annotation and evaluation.
% The dataset containing 1905 video clips for a herd of 200 Holstein-Friesian cows will be made publicly available at \href{https://github.com/SolanLiu25/Dataset_CattleBehaviours6}{CattleBehaviours6}, together with standardized training, validation, and test splits for benchmarking purposes.

%%%%%%%%%%%%%%%%%%%%%%%%%%%%%%%%%%%%%%%%%%
\section{{Materials---The} %\textcolor{( We revised the hyphen to emdash there, please confirm.)}
 {CattleBehaviours6} %\textcolor{( Please confirm if the italic is unnecessary and can be removed.)}
 Dataset}
\label{dataset curation}

\subsection{Behaviour Definitions}
\label{subsec:behdef}
Prior to the collection and annotation of cattle data, it is essential to establish an ethogram outlining the range of scientifically relevant behaviours along with their precise definitions. This is particularly important when the data annotators are not specialist ethologists, as clear guidelines help ensure consistency and reduce subjective interpretation during the labelling process.
Currently, there is no universally standardised ethogram for cattle behaviour. While numerous ethograms tailored for observational veterinary research exist in the literature \citep{li2024dynamic, han2024beef}, they often require contextual cues that are currently only available to human observers and thus do not necessarily map to computer vision-based behaviour analysis.
In practice, behaviours automatically analysed by computational models currently need to be simplified. For example, while a human observer can easily use contextual cues, such as the source of food, to distinguish the chewing movement as part of \textit{ruminating} or \textit{feeding}, computer vision models often face limitations due to restricted video viewports and short clip durations.
For this reason, prior studies in computer vision \citep{fuentes2020deep, qiao2022c3d, geng2024motion} have developed task-specific ethograms adapted to their particular research goals and data modalities. 
\re{Here, an ethogram focused on six basic individual behaviours is defined, including \textit{feeding}, \textit{drinking and posture-dependent states of self-grooming and ruminating}, which differ in the visibility and temporal patterns of their observable cues.
Both \textit{standing-ruminating} and \textit{lying-ruminating} are included because rumination can occur in different postural states, which result in distinct visual appearances in video.}

To ensure behavioural clarity and annotation consistency, the behavioural definitions were established according to the following principles.
{Each} %\textcolor{( Please confirm if the bold of paragraph is unnecessary and can be removed, please check and confirm all bold in whole tex. )}
 behaviour follows a standardised description template: \textbf{``The focal cow \{action name\}, \{attributes\}''}. Attributes may consist of the key body part involved, the context, the cows' posture (standing or lying) and the temporal characteristics (i.e., whether it constitutes as states or events {(Events are instantaneous; states have appreciable durations \citep{altmann1974observational})}%\textcolor{( Footnote is not permitted in this journal, so we have moved it into the text, please confirm the whole tex. )}
). Although all annotated behaviours in the current dataset are categorised as states, the conceptual distinction between states and events remains important, particularly for future expansion of the dataset to include instantaneous behaviours. Additionally, to reduce complexity and ambiguity, non-exclusive behaviours (i.e., behaviours that may co-occur or overlap in time) were excluded.
{Table~\ref{tbl1} shows detailed definitions of the six commonly observed behaviours, together with \textit{standing} and \textit{lying} as auxiliary posture definitions used to describe posture-dependent behaviours.}

\begin{table}[H]
    \caption{{Cattle} %\textcolor{(1. We revised top alignment to middle in the table, please confirm. 2. We added table lines to make the table content to be shown more clear in the table, please confirm. 3. Please note that the size of figures and tables, the position of figures and tables will be adjusted appropriately may occur during the production stage to reduce the page blank. )}
 behaviour definitions. Each behaviour is defined by characteristic movement patterns, typical body part involvement, or locations. All behaviours presented here are classified as behavioural states.\label{tbl1}}
    \begin{adjustwidth}{-\extralength}{0cm}
   % \centering
    \small
    \newcolumntype{L}[1]{>{\raggedright\arraybackslash}m{#1}}
\newcolumntype{C}[1]{>{\centering\arraybackslash}m{#1}}
\newcolumntype{R}[1]{>{\raggedleft\arraybackslash}m{#1}}
\begin{tabular}{L{3.5cm}L{12cm}L{1.71cm}}
%    \begingroup
%    \renewcommand{\tabularxcolumn}[1]{>{\raggedright\arraybackslash}p{#1}}
%    
%    \begin{tabularx}{\fulllength}{@{}
%    >{\raggedright\arraybackslash}p{0.20\fulllength}
%    >{\raggedright\arraybackslash}X
%    >{\raggedright\arraybackslash}p{0.12\fulllength}
%    @{}}
    \toprule
    \textbf{Behaviour} & \textbf{Description} & \textbf{Posture}\\
    \midrule
    
    standing & The focal cow supports its body weight with all four limbs on the ground directly below the body. & standing \\
    
   \bottomrule
    \end{tabular}
   % \endgroup
    
    \end{adjustwidth}
\end{table}

\begin{table}[H]\ContinuedFloat
\caption{{\em Cont.}}
    \label{tbl1}
    \begin{adjustwidth}{-\extralength}{0cm}
   % \centering
    \small
    \newcolumntype{L}[1]{>{\raggedright\arraybackslash}m{#1}}
\newcolumntype{C}[1]{>{\centering\arraybackslash}m{#1}}
\newcolumntype{R}[1]{>{\raggedleft\arraybackslash}m{#1}}
\begin{tabular}{L{3.5cm}L{12cm}L{1.71cm}}
%    \begingroup
%    \renewcommand{\tabularxcolumn}[1]{>{\raggedright\arraybackslash}p{#1}}
%    
%    \begin{tabularx}{\fulllength}{@{}
%    >{\raggedright\arraybackslash}p{0.20\fulllength}
%    >{\raggedright\arraybackslash}X
%    >{\raggedright\arraybackslash}p{0.12\fulllength}
%    @{}}
    \toprule
    \textbf{Behaviour} & \textbf{Description} & \textbf{Posture}\\

    \midrule
    
    lying & The focal cow supports its body weight with the sternum in contact with the ground and the limbs can be stretched away from the body or tucked underneath the body. & lying \\
    \midrule
    feeding & The focal cow has its head through the feed rails and inside the feeding trough/area. & standing \\\midrule
    
    drinking & The focal cow has its head inside or over the water trough. & standing \\\midrule
    
    standing-self-grooming & The focal cow while standing uses its tongue to lick parts of its body. This can also be seen as mouth movement in contact with the body. Excluding licking nostrils (or nose cleaning) as well as scratching against farm equipment or bedding. & standing \\\midrule
    
    standing-ruminating & The focal cow while standing regurgitates a bolus (ball of feed from the rumen) followed by chewing in a rhythmic motion and swallowing of the bolus. \re{This cycle of regurgitation, chewing and swallowing may repeat during the observed \mbox{rumination behaviour.}} & standing \\\midrule
    
    lying-self-grooming & The focal cow while lying uses its tongue to lick parts of its body. This can also be seen as mouth movement in contact with the body. Excluding licking nostrils (or nose cleaning) as well as scratching against farm equipment or bedding. & lying \\\midrule
    
    lying-ruminating & The focal cow while lying regurgitates a bolus (ball of feed from the rumen) followed by chewing in a rhythmic motion and swallowing of the bolus. \re{This cycle of regurgitation, chewing and swallowing may repeat during the observed rumination behaviour.} & lying \\
    
    \bottomrule
    \end{tabular}
   % \endgroup
    
    \end{adjustwidth}
\end{table}

\subsection{Data Collection}
The video footage was captured between {27 October 2024} %\textcolor{( We revised the date format, please confirm the revisions in whole tex.)}
 and 20 April 2025 at the John Oldacre Centre, Bristol Veterinary School, UK, a 60-camera research platform for intensive monitoring. The experimental platform was established within an operational commercial dairy farm comprising approximately 200 Holstein Friesian cows. 
\re{Milking cows were housed in a free-stall barn with sand-bedded cubicles and ad libitum access to feed. The milking herd was divided into two groups: Group 1, with a maximum of 92 cows, comprised cows in lactations 1--3, while Group 2, with approximately 108 cows, comprised cows in lactation 3 and above. The herd was milked three times daily, from \mbox{06:00--08:00,} 14:00--16:00 and 22:00--00:00. Transition cows, defined as cows within three weeks of calving, were housed separately in a loose-straw-bedded pen. Individual health status was not collected as part of this study. Environmental variables such as barn temperature and humidity were not routinely recorded because this study was conducted within a commercially managed dairy herd.
Recordings were conducted over 20 non-consecutive recording days between October 2024 and April 2025. 
The specific recording dates are summarised in Table~\ref{tab:recording_dates} in Appendix~\ref{app:schedule}.}
Cameras were ceiling-mounted approximately four metres above the ground and angled to provide clear views of the feed face or resting areas where applicable. Videos from two types of cameras ({Hanwha QNV-C8012, Hanwha Vision, Seoul, South Korea} %\textcolor{( Please state the name of the manufacturer, city, and country from where the equipment was sourced. Same as below.)}
 and {Hikvision DS-2CD5546G0-IZHSx, Hikvision, Hangzhou, China}) were used in this study.

Although the precise number of individual cows observed across the clips is uncertain, efforts were made to ensure broad coverage by using seven non-overlapping camera views and conducting data collection over several days.
To enrich the diversity of the collected data, videos were collected in varying weather conditions (rainy, cloudy, sunny) and at different times of the day (early morning, morning, noon, afternoon, late afternoon, night-time with artificial lighting), allowing for a wide range of illumination conditions (See Figure~\ref{FIG:1}a). 
Dim conditions, common in early mornings, late afternoons and nighttime scenes, frequently obscured parts of the cows in darkness, reducing the visibility of key features. In contrast, strong illumination typically occurred during sunny mornings and afternoons, which often smoothed the surface textures of the cows due to over-exposure. 
Multiple cameras were also employed in different functional areas, including feeding, drinking and resting zones to introduce variation in camera perspectives and background environment (See Figure~\ref{FIG:1}b).

\begin{figure}[H]

    \subfloat[\centering Illumination\_Variety.]{
        \includegraphics[width=0.99\textwidth]{Illumination\_Variety.png}
    }
    
  %  \vspace{3mm}
    
    \subfloat[\centering Camera\_views.]{
        \includegraphics[width=0.99\textwidth]{Camera\_views.png}
    }
    
    \caption{{Data} %\textcolor{( 1. We removed large spaces and put label a, b to center format in figure body, please confirm. 2. Please confirm whether the overlapping/incomplete content in this figure affects scientific understanding and if it does, please revise it. 3. Without affecting refs. citations or page blank, it is recommended to move figures and tables in the text to after their first mention during the layout stage. Please confirm the location revision of some figures.)}
 variability. (\textbf{a}) Illumination variability across the dataset includes ``night'' scenes with artificial lighting and ``dim'' conditions with natural lighting, both of which cause parts of the cows to be obscured in darkness. In contrast, ``strong light'' conditions tend to smooth the cows' surface textures. (\textbf{b}) The dataset was collected using seven non-overlapping camera views. Each camera's number is shown in red. All behaviours were observable across cameras, except \textit{feeding}, which was not visible in cameras $4$ and $5$.
    \label{FIG:1}}
\end{figure}

Video resolution ranged from $1280\times720$ to $2560\times1440$ pixels, and the frame rate varied between $16$ and $25$ frames per second (FPS). This variation is due to data being collected over different periods, including a camera configuration phase that took place in the winter of 2024.

\subsection{Data Annotation}
An annotation workflow was developed to minimise the need for manual intervention. As a preprocessing step, the original videos were downsampled to 4 or 5 FPS and resized to half of their original resolution, in order to reduce storage requirements while still maintaining sufficient spatial and temporal resolution for reliable behaviour observation. The preprocessed videos then served as the input to the subsequent cattle detection and tracking stages.

To develop the cattle detection algorithm, T-Rex Label \citep{jiang2024t} was used to annotate the position of cows automatically with minimal manual intervention. This process yielded $1759$, $219$ and $219$ images for the training, validation and testing dataset, respectively. For model training, YOLOv11 \citep{khanam2024yolov11} was applied due to its fast inference speed. The model achieved an $mAP_{50}$ of $98.2\%$ on the test data. 
To obtain tracking identifiers and movement trajectories for individual cows, ByteTrack \citep{zhang2022bytetrack} was incorporated into the framework.

Using the bounding boxes obtained from the detector and also the tracklets obtained from the tracker (See Figure~\ref{FIG:2}), fixed-length video clips of 50 frames were extracted from the full-length recordings, ensuring each clip ideally contained one central cow. All frames within each clip were kept at the same resolution and aspect ratio. However, isolating single-cow scenes proved challenging due to the frequent spatial and temporal occlusions caused by limited camera views, overlapping cows and individual cow movement. 
To maintain dataset consistency, specific rules were introduced to handle scenarios with multiple cows and occlusions:

\begin{enumerate}
  \item For instances containing multiple cows, common during feeding time, clips were retained only when all visible cows (typically their heads) were engaged in the \mbox{same behaviour.} 
  \item In the case of spatial occlusion, the central cow was required to occupy at least half of the image area. 
  \item For temporal occlusion, the target behaviour had to be observable for at least two-thirds of the clip duration.  
\end{enumerate}
\vspace{-3pt}

\begin{figure}[H]
	%\centering
    \includegraphics[width=.95\textwidth]{Annotation\_flow.png}
    \vspace{-6pt}
    \caption{{Automatically} %\textcolor{( 1. Please confirm whether the overlapping/incomplete content in this figure affects scientific understanding and if it does, please revise it. 2. Please confirm whether an explanation of the arrow needs to be added to the figure caption. 3. We inserted it AFTER where it is first mentioned in the text, please check and confirm. )}
 generated fixed-length clips. The clips were extracted using the cattle detector and tracker developed, allowing consistent sampling around individual cows for downstream behaviour analysis. Red boxes indicate detection results, while green solid lines represent the tracking trajectories (tracklets).}
	\label{FIG:2}
\end{figure}

Based on these criteria, the spatial crop or clip length was occasionally adjusted to maximise the amount of usable data retained. All selected clips were manually labelled according to the behavioural definitions described in Section~\ref{subsec:behdef}. 
{The annotations were produced through a two-stage process involving two annotators with backgrounds in computer science and animal science. First, the annotator with a computer science background performed the initial annotation under the guidance of the animal science annotator. This stage included refinement of the operational behaviour definitions and discussion of potentially ambiguous cases. Second, the completed annotations were reviewed by the animal science annotator, and any ambiguous cases were discussed before the final labels were confirmed. As this procedure was designed as an expert-guided annotation and review process rather than independent duplicate annotation, formal inter-rater agreement was not estimated.}
Several points warrant clarification. Firstly, video clips may originate from the same behaviour bout of an individual cow, since behavioural movements and background conditions may change gradually within a bout while still introducing sufficient visual variation across consecutive clips. \re{Secondly, \textit{feeding} and \textit{drinking} were defined using observable positional and contextual cues rather than direct confirmation of active ingestion. Future work will consider finer-grained definitions of active feeding and drinking. }
Likewise, some clips labelled as \textit{standing-ruminating} may require additional contextual information to further determine whether the behaviour represents true rumination or is part of feeding. 
This reflects a practical challenge related to the annotation in distinguishing behaviours solely from visible movements without contextual cues. In particular, when observing short video clips presented in random order, it is difficult to reliably infer the cow's location or access to food, both of which are crucial for accurate recognition in ethology. 

The resulting dataset contains 1905 clips (approximately five hours of footage), each labelled with one of the six basic behaviours, some examples of which are shown in Figure~\ref{FIG:3}. The clip durations range from three to ten seconds at a fixed rate of 5 FPS. The dataset was randomly partitioned at the clip level into training, validation and test sets using a 6:2:2 split, without considering class balance. 
\re{Because persistent animal identities were unavailable, the partition was not performed at the individual-cow level.
In addition, a temporally held-out test set comprising 351 clips collected on 20--21 February 2025 was kept separate from model training and validation and used only for additional evaluation.}
\vspace{-3pt}
\begin{figure}[H]
  %  \centering
    \includegraphics[width=.99\textwidth]{Data\_map-v2.png}
    \caption{Illustration of the six behaviours used for annotation. For each behaviour category, multiple representative clips are presented to illustrate intra-class variability in appearance, lighting and camera viewpoints. Numbers refer to the number of labelled instances per category.}
	\label{FIG:3}
\end{figure}

%%%%%%%%%%%%%%%%%%%%%%%%%%%%%%%%%%%%%%%%%%
\section{Methodology}
To contextualise the proposed approach, the underlying classification mechanism of CLIP and its extension to video are first reviewed before describing our Cattle-\mbox{CLIP approach.}

\subsection{Preliminary}
The classification mechanism of CLIP is first reviewed to provide context for the proposed approach. CLIP contains two parallel Transformers \citep{dosovitskiy2020image}, denoted by ${f_{img}, f_{text}}$ to generate image and text representations, respectively:

\textbf{Image encoding}. Given an image sample $I \in \mathbb{R}^{H\times W\times 3}$ of spatial resolution $H \times W$, the image is first divided into non-overlapping $P \times P$ patches and flattened to a set of vectors $\{x_{j} \in \mathbb{R}^{N \times 3P^2}\}_{j=1}^N$, where $j$ is the patch index, and the number of patches $N=HW/P^2$. An image-level token sequence $i_j^0 = Conv(x_j)$ is obtained by applying a shared convolutional projection to all patches, with the superscript $0$ denoting the initial input to the Transformer prior to encoding. The input sequence is prefixed with a learnable classification embedding, known as the [CLS] token $x_{cls}$, followed by the addition of a positional embedding $e_{pos}$. The resulting input token sequence $\hat{i}^0$ is defined {as:} %\textcolor{( Please carefully check variables formatting (italic, bold, subscript, uppercase, etc.) throughout the manuscript to ensure the formatting is consistent and revise if needed. )}
\begin{equation*}
  \hat{i}^0 = [x_{cls}, i^0]+e_{pos}.
\end{equation*}

{Within} %\textcolor{( Please confirm if the noindent of paragraph is necessary in whole tex.)}
 the vision encoder ${f_{img}}$, $\hat{i}^{l}$ is progressively encoded at each Transformer layer $l \in \{ 1, \cdots, L_{img}\}$, given by:
\begin{equation*}
  \hat{i}^l = f_{img}^l(\hat{i}^{l-1}).
\end{equation*}

The [CLS] token from the last layer $\hat{i}_{cls}^{L_{img}}$ is projected onto a shared vision--language latent space using a matrix $A_{img}$, yielding an image-level representation $i = \hat{i}_{cls}^{L_{img}} A_{img}$.

\textbf{Text encoding}. Currently, the input $t^0$ to the text encoder refers to the sequence of embedded token vectors derived from category descriptions. To better align the textual and visual modality, each category description is formatted within a natural language prompt template, such as ``\textit{a photo of a <category>}{.}%\textcolor{(  Please confirm if the period is necessary there.)}
'', prior to being tokenised and embedded.  
An end-of-sequence [EOS] token is appended to mark the end of the sentence. The embedded input is then passed through the text encoder $f_{text}$, which performs sequential processing over Transformer layers $l \in \{ 1, \cdots, L_{text}\}$ as follows:
\begin{equation*}
  t^l = f_{text}^l(t^{l-1}).
\end{equation*}

Similarly, the final text embedding $t$ is computed by applying a projection matrix $A_{text}$ to the [EOS] token from the final Transformer layer, mapping it into the shared vision–language latent space:
\begin{equation*}
  t = t_{eos}^{L_{text}} A_{text}.
\end{equation*}

\textbf{Image--text matching}. During training, the model optimises the cosine similarity, $sim(\cdot)$, between the image-level embeddings, $i$, and their corresponding text embeddings, $t$, across a batch. This similarity is maximised using a contrastive cross-entropy loss with a temperature parameter $\tau$:
\begin{equation*}
  \mathcal{L} = -\mathbb{E}_{(i,t)} \log \frac{\exp \frac{(sim(i, t))}{\tau}}{{\sum\limits_{t \in \mathcal{T}} \exp(\frac{sim(i, t)}{\tau})}},
\end{equation*}
where $\mathcal{T}$ denotes the collection of all class-level textual descriptions. The cosine similarity function is defined as:
\begin{equation*}
sim(i,t) = \frac{i\cdot t}{||i|| ||t||}.
\end{equation*}

\textbf{Extending CLIP for video understanding.} 
Unlike vanilla CLIP, which is designed for static image--text alignment, our task needs to extend the image encoder to process video inputs. Following prior works on adapting CLIP to video tasks ~\citep{wang2021actionclip, rasheed2023fine, wu2023revisiting}, temporal integration was incorporated over frame-level features before the video–text similarity calculation to produce a unified video representation. Specifically, given a sequence of 
$K$ sampled frames from an input video, features were extracted for each frame independently using the CLIP image encoder and then aggregated via average pooling:
\begin{equation*}
  v = AvgPool([i_1, \cdots , i_{K}]).
\end{equation*}

\subsection{Cattle-CLIP}
Cattle-CLIP is introduced as a multimodal framework for automated cattle behaviour recognition that integrates customised augmentation and domain-specific text prompts to embed domain knowledge into the learning process:

\textbf{Customised augmentation}. In general, random crops are commonly used in classification tasks to augment image datasets and increase their diversity. However, this strategy is not well suited to our dataset. Preserving the visibility of critical features, particularly the head and mouth movement of the central cow, is essential for accurately capturing cows' behaviour-related features. As shown in Figure~\ref{FIG:5a}, the random \textit{crop} operation results in the loss of head information, leading to an incomplete appearance. Therefore, the \textit{crop} augmentation was removed to avoid unintentionally truncating these informative regions.
Moreover, the frames in our dataset exhibit a notably wide range of aspect ratios (as illustrated in Figure~\ref{FIG:5b}), which results in significant information loss during standard cropping. To address this issue, the \textit{crop} operation was removed to preserve frame integrity and instead \textit{fill} the frames to maintain the original aspect ratio before resizing them to the target resolution. The constant-value padding technique was employed to standardise each of the sampled frames as a square. This approach preserves both the content and spatial consistency of the original frames across each clip and leads to clear gains in model performance. After \textit{resizing}, additional augmentations such as flipping, colour jittering and grayscale transformations were utilised, consistent with methods used in other research~\citep{rasheed2023fine}. This augmentation pipeline aims to enhance the robustness of the model while maintaining the critical features necessary for behaviour analysis.

\begin{figure}[H]
%	\centering
      \includegraphics[width=.85\textwidth]{Data\_augmentation.png}
      \vspace{-3pt}
    \caption{{Comparison} %\textcolor{(  Please confirm whether an explanation of the arrow needs to be added to the figure caption.)}
 of original images and augmented versions using two different augmentation strategies. The commonly used \textit{crop} operation often removes key body parts of cows. In contrast, applying \textit{fill} before \textit{resize} better preserves core behavioural cues, such as head movements important for most of the behaviour recognition, and maintains the structural integrity of the cows.} 
    \label{FIG:5a}
\end{figure}

\vspace{-9pt}

\begin{figure}[H]
%	\centering
    \includegraphics[width=.5\textwidth]{Ratio\_statistics.png}
    \vspace{-5pt}
    \caption{Histogram of video aspect ratios after cropping. The dataset contains videos with a wide range of aspect ratios, posing challenges for standard \textit{crop} operations, motivating aspect-ratio-aware processing strategies.}
    \label{FIG:5b}
\end{figure}

\begin{figure}[H]
	%\centering
    \includegraphics[width=.99\textwidth]{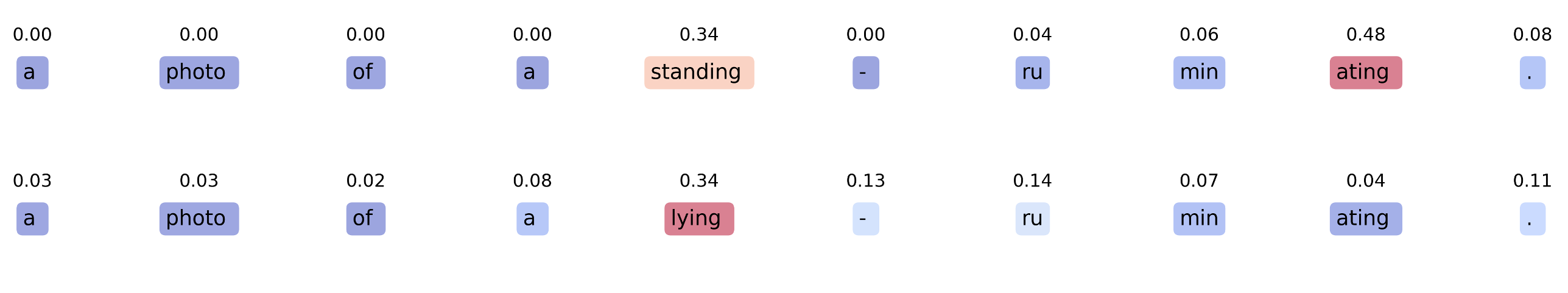}
    % \vspace{-5mm}
    \caption{{Attention} %\textcolor{( Please confirm if the hyphen is correct and if it needs to add an explanation in figure caption.)}
 map for the text tokens of \textit{ruminating}. The numerical values indicate the level of attention assigned to each token.}
    \label{FIG:6}
\end{figure}

\textbf{Domain-specific text prompts}.
Following the findings of CLIP, which demonstrate that using full-sentence prompts yields better performance than a single word, the default text template ~``a photo of a <category>.'' was adopted as our input prompt. 
A more grammatically complete variant, ``a photo of a cow <category>,'' was also evaluated; however, it did not improve performance.
However, a gap was observed in the language used to depict human activities and animal behaviours. Upon visualising the attention \mbox{map \citep{chefer2021generic}} on the text (shown in Figure~\ref{FIG:6}), it is seen that the CLIP model with default text prompts splits the word ``\textit{ruminating}'' into three separated subword tokens: \textbf{ru}, \textbf{min} and \textbf{ating}. This division disrupts the semantic meaning of the class name. 
To address this, the text prompt was modified by replacing ``\textit{ruminating}'' with a more interpretable phrase that could be better recognised by CLIP. Specifically, the phrase ``\textit{chewing}'' from the behaviour definition was adopted to facilitate the model's ability to capture the intended semantics. For other behaviours, the original class names were retained since similar tokenisation issues were not observed. This adjustment not only improved the model's performance but also aligned the prompts with more meaningful and consistent representations of \mbox{animal behaviours.}

%%%%%%%%%%%%%%%%%%%%%%%%%%%%%%%%%%%%%%%%%%
\section{Experiments}

\subsection{Fully Supervised Scenario}
Here, the entire dataset is annotated with class labels, and the total number of classes is predefined and known during training. In this setting, models are trained on all training examples and evaluated on the corresponding test set.

\subsection{Few-Shot Scenario}
To evaluate Cattle-CLIP under low-data conditions for data-limited behaviours, a two-stage base-to-novel few-shot learning strategy was adopted. In real-world applications, such data scarcity is common, as certain behaviours occur infrequently or correspond to newly introduced categories during dataset expansion.
In this strategy, the first stage trains the model on five selected behaviours, referred to as the \textit{base} behaviours, using the base dataset $D_b$ and evaluates on the same set of behaviours. The remaining behaviour is treated as the \textit{data-scarce} behaviour, with only a limited number $n$ of samples available, where $n \in \{16, 8, 4, 2 \}$. These samples constitute the data-scarce dataset $D_r$.

To mitigate the well-known issue of catastrophic forgetting in transfer learning, a replay strategy was adopted that incorporates both the data-scarce dataset and a randomly sampled subset of the base dataset. Specifically, a new training dataset \mbox{$D_f = D_r \cup sample(D_b)$} was constructed, consisting of all available samples from $D_r$ and a comparable number of samples from $D_b$. This balanced dataset $D_f$ is then used in the second stage to train a final classifier $F$, enabling the model to retain previously learned knowledge while adapting to the novel classes. The final model $F$ is evaluated on the full test set ($381$ clips) with the expectation of recognising all behaviours.
To ensure robustness, three different splits of $D_f$ are generated using different random seeds.

This process is repeated for each of the six behaviours in a leave-one-out manner, treating each one as the data-scarce behaviour in turn. As a result, six base classifiers were obtained, denoted as $B_{feed},$ $B_{drink},$ $B_{std\_groom},$ $B_{std\_rumi},$ $B_{ly\_groom},$ $B_{ly\_rumi}$ and a series of final models $F_{feed}^n,$ $F_{drink}^n,$ $F_{std\_groom}^n,$ $F_{std\_rumi}^n,$ $F_{ly\_groom}^n,$ $F_{ly\_rumi}^n$. 

\textbf{Baselines}. For comparison, a series of one-stage baseline models $M_{feed}^n,$ $M_{drink}^n$, $ M_{std\_groom}^n,$ $M_{std\_rumi}^n, $ $M_{ly\_groom}^n,$ $M_{ly\_rumi}^n$ were each trained on five base behaviours and one data-scarce behaviour using the combined dataset $D_m = D_r \cup D_b$. Importantly, the $D_r$ used here is exactly the same as that employed in the final classifiers, ensuring that performance differences arise from the training approach rather than differences in data subset. These baseline models are used to assess the Cattle-CLIP's ability to learn under imbalanced and low-resource conditions in a single-stage setting.

\subsection{Common Settings}
The ViT-B/16 variant of the CLIP model was adopted for our experiments. Specifically, ViT-B/16 refers to the base configuration of the Vision Transformer (ViT), which consists of $L=12$ Transformer layers, a hidden size of $768$ and a patch size of $P=16$ pixels. The non-overlapping $16 \time 16$ patches are then linearly projected into tokens before being fed into \mbox{the encoder.}

All experiments were conducted using a single NVIDIA GeForce RTX 2080 Ti GPU with 11 GB of memory for training. 
For video sampling, $K = 8$ frames were randomly selected from each video clip, a commonly used setting in video action recognition benchmarks (alongside $16$ and $32$). Although using more frames could potentially capture richer motion information, $K = 8$ was selected due to training-time memory constraints, which limited the feasible batch size and model complexity.
Importantly, this random sampling is repeated at the start of each training epoch, meaning that the model is exposed to different subsets of frames across epochs.
Following CLIP, the selected frames are then preprocessed to a fixed spatial resolution of $H \times W = 224 \times 224$. For the textual modality, the maximum sequence length is constrained to 77 tokens, including special [EOS] token. 
The AdamW optimiser \citep{loshchilov2017decoupled} was employed with a weight decay of $1 \times 10^{-3}$, and a cosine learning rate scheduler with an initial learning rate of $2.2 \times 10^{-5}$ was adopted. 
The default training configuration uses a batch size of 4 with a gradient accumulation step of 2 to effectively simulate a larger batch size under memory constraints. 
Unless otherwise specified, models are trained for 30~epochs with a 5-epoch warm-up schedule. 

To accommodate different training paradigms, the initialisation scheme, initial learning rate and total number of epochs were adjusted accordingly.
Specifically, the $M^n$ and $B$ models were initialised with pretrained CLIP parameters and fully fine-tuned on all model parameters. These models are trained for 30 epochs in total. The final models $F^n$, trained under the few-shot learning setting, were initialised with the weights of the corresponding $B$ models and then fully fine-tuned. Among them, $F^{16}$ models are trained for 30 epochs (the same as $M^n$ and $B$), while $F^{2}, F^{4}$ and $F^8$ models are trained for 100 epochs to allow repeated exposure to the limited data. The initial learning rate for $F^n$ models is set to $2.2 \times 10^{-4}$, whereas a lower rate of $2.2 \times 10^{-5}$ is used for both $M^n$ and $B$ models.

{To evaluate model performance, accuracy, confusion matrices, per-class precision, recall and F1-score were used as the primary evaluation metrics. For each class, precision and recall were calculated from the numbers of true positives (TP), false positives (FP) and false negatives (FN) as
\begin{equation*}
\mathrm{Precision} = \frac{TP}{TP + FP}, \mathrm{Recall} = \frac{TP}{TP + FN}.
\end{equation*}

The F1-score was calculated as the harmonic mean of precision and recall:
\begin{equation*}
F_1 = 2 \times \frac{\mathrm{Precision} \times \mathrm{Recall}}
{\mathrm{Precision} + \mathrm{Recall}}.
\end{equation*}

Confusion matrices were generated by comparing the predicted class label with the corresponding ground-truth label for each sample in the test set. Rows represent the ground-truth classes, whereas columns represent the predicted classes. Each cell reports the number of samples from a given ground-truth class that were assigned to the corresponding predicted class. Per-class recall is shown on the right-hand side of the matrix, while per-class precision is shown below the matrix.}

%%%%%%%%%%%%%%%%%%%%%%%%%%%%%%%%%%%%%%%%%%
\section{Results and Analysis}
\subsection{Fully Supervised Evaluation}
Cattle-CLIP achieves an overall accuracy of 96.1\% across all behaviour categories. As shown in the confusion matrix in Figure~\ref{FIG:111} and per-class F1-score in Table \ref{tab:f1_score}, the model performs exceptionally well on \textit{feeding} and \textit{drinking} behaviours, achieving precision and recall scores of 0.97/1.00 and 0.98/0.98, respectively. However, it struggles somewhat to distinguish between \textit{lying-self-grooming} and \textit{lying-ruminating}, likely due to its limited ability to capture fine-grained temporal features.

\begin{figure}[H]
	%\centering
    \includegraphics[width=.55\textwidth]{Cattle-CLIP\_cm.png}
    \caption{{Confusion} %\textcolor{( Please confirm whether the explanations of the red font and colors need to be added to the figure caption.)}
 matrix of Cattle-CLIP on CattleBehaviours6 test set. The model demonstrates category-dependent performance, achieving high accuracy in recognising \textit{feeding} and \textit{drinking} behaviours; however, it exhibits confusion between \textit{lying-self-grooming} and \textit{lying-ruminating} behaviours.}
	\label{FIG:111}
\end{figure}

\vspace{-9pt}

\begin{table}[H]
    \caption{Per-class F1-score on CattleBehaviours6 test set under the fully supervised setting.\label{tab:f1_score}}
    \begin{adjustwidth}{-\extralength}{0cm}
 %   \centering
%    \small
    \begin{tabularx}{\fulllength}{@{}CccCCCC@{}}
    \toprule
    \textbf{Behaviours} & \textbf{Feeding} & \textbf{Drinking} & \textbf{Standing-Self-Grooming} & \textbf{Standing-Ruminating} & \textbf{Lying-Self-Grooming} & \textbf{Lying-Ruminating} \\
    \midrule
    F1-score & 0.99 & 0.98 & 0.95 & 0.97 & 0.89 & 0.94 \\
    \bottomrule
    \end{tabularx}
    \end{adjustwidth}
\end{table}

{To further assess robustness under different recording conditions, two supplementary evaluations were conducted. First, a camera-specific evaluation showed comparable performance across the two camera groups, with overall accuracies of approximately 95.5\% and 96.7\%, respectively. Detailed confusion matrices are provided in Appendix~\ref{app:camera}. Because camera type was associated with fixed recording locations, these results should be interpreted as a camera-group comparison rather than a controlled assessment of camera hardware alone.
Second, Cattle-CLIP was evaluated on the temporally held-out test set described in Section~\ref{dataset curation}.
The behaviour distribution of this additional test set is summarised in Table~\ref{tab:additional_distribution}. 
On this dataset, the model achieved an overall accuracy of 94.6\%. This slight performance variation compared with the main evaluation results (96.1\%) is considered acceptable and expected, given the natural environmental variations between recording sessions.
The corresponding confusion matrix is presented in Figure~\ref{fig:additional_confusion}.}

\begin{table}[H]
    \caption{Data distribution and per-class F1-score of temporally held-out test set.\label{tab:additional_distribution}}
    \begin{adjustwidth}{-\extralength}{0cm}
   % \centering
    \begin{tabularx}{\fulllength}{@{}lCCCCCC@{}}
    \toprule
    \textbf{Behaviours} & \textbf{Feeding} & \textbf{Drinking} & \textbf{Standing-Self-Grooming} & \textbf{Standing-Ruminating} & \textbf{Lying-Self-Grooming} & \textbf{Lying-Ruminating} \\
    \midrule
    Number & 77 & 48 & 53 & 54 & 36 & 83 \\ 
    \midrule
    F1-score & 0.99 & 0.95 & 0.95 & 0.95 & 0.86 & 0.93 \\
    \bottomrule
    \end{tabularx}
    \end{adjustwidth}
\end{table}

\vspace{-6pt}

\begin{figure}[H]
	%\centering
    \includegraphics[width=.55\textwidth]{Cattle-CLIP\_cm_add.png}
    \caption{{Confusion} %\textcolor{( Please confirm whether the explanations of the red font and colors need to be added to the figure caption.)}
 matrix of Cattle-CLIP on the temporally held-out test set.}
	\label{fig:additional_confusion}
\end{figure}
  
\subsection{Comparison with State of the Art}
Cattle-CLIP was compared with several state-of-the-art video-based methods. Specifically, Video Swin \citep{liu2022video}, X-CLIP \citep{ni2022expanding} and EVL \citep{rasheed2023fine} were implemented using the default optimal hyperparameters reported in their respective studies. The comparison results are presented in Table \ref{tbl2}. As shown, Cattle-CLIP achieves the highest overall accuracy, demonstrating its effectiveness relative to existing approaches.

\begin{table}[H]
    \caption{Comparison of model accuracy. The vanilla model is CLIP with a temporal integration layer. Here, ``+aug'' indicates adding customised augmentation to the vanilla model, while ``+textPrompts'' refers to replacing the original default text prompts with model-understandable words.
    \label{tbl2}}
   % \centering
    \small
    \begin{tabularx}{\textwidth}{lC}
    \toprule
    \textbf{Models} & \textbf{Accuracy (\%)} \\
    \midrule
    Video-Swin (base) \cite{liu2022video} & 93.2 \\
    X-CLIP (ViT-B/16) \cite{ni2022expanding} & 93.9 \\
    EVL (ViT-B/16) \cite{rasheed2023fine} & 91.1 \\
    
    \midrule
    Vanilla & 93.4 \\
    Vanilla + aug & 95.5 \\
    Vanilla + text prompts & 94.2 \\
    Cattle-CLIP (Vanilla + aug + text prompts) & \textbf{{96.1} %\textcolor{( Please add an explanation for the use of bold in the table footer. If the bold is unnecessary, please remove it. )}
} \\
    \bottomrule
    \end{tabularx}
\end{table}

\subsection{Ablation Study}
To provide additional verification of our approach, a vanilla model was included, adding only the temporal integration layer after the original vision encoder. The proposed components, including domain-specific augmentation and text prompts, were then incorporated incrementally.
The results, presented in Table \ref{tbl2}, demonstrate that both the \textit{fill} augmentation strategy and the use of semantically understandable prompts, tailored for animal behaviours, play a critical role in minimising the domain gap, yielding performance gains of $2.2\%$ and $0.9\%$ over the vanilla model. When combined, these two strategies result in a total performance increase of $2.9\%$.
  
\subsection{Few-Shot Evaluation}
\label{sec: 5}
\begin{figure}[H]%[!thbp]
	%\centering
  \includegraphics[width=.98\textwidth]{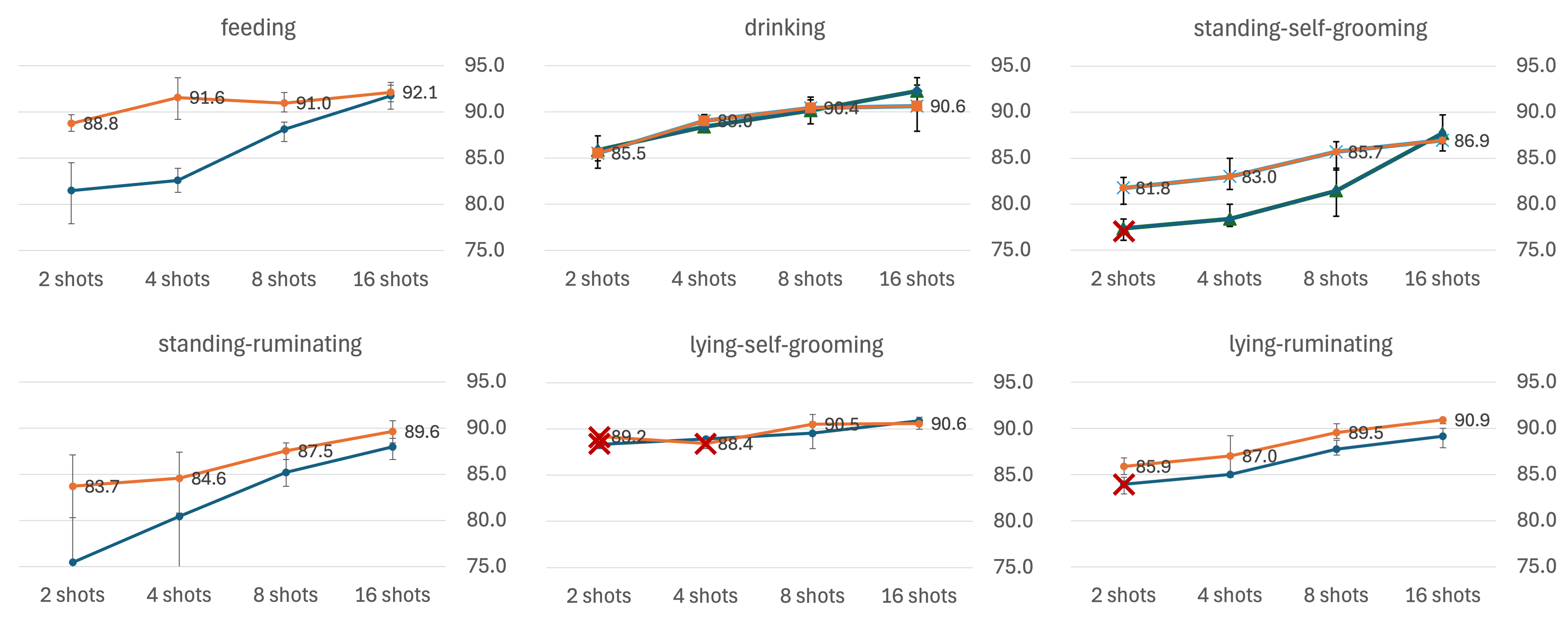}
  % \vspace{0.5mm}
  \caption{{Mean} %\textcolor{( Please confirm whether the overlapping/incomplete content in this figure affects scientific understanding and if it does, please revise it.)}
 accuracy (\%) curves with minimum and maximum error bars over three dataset splits for different learning paradigms. The figure compares the performance of two learning paradigms: (1)~orange lines: the proposed two-stage base-to-novel few-shot learning models, and (2) blue lines: the one-stage baseline models. The models, arranged from left to right and top to bottom, are denoted as $\{ (F_{feed}^n, M_{feed}^n),$ $(F_{drink}^n, M_{drink}^n),$ $(F_{std\_groom}^n, M_{std\_groom}^n),$ $(F_{std\_rumi}^n, M_{std\_rumi}^n),$ $(F_{ly\_groom}^n, M_{ly\_groom}^n),$ $(F_{ly\_rumi}^n, M_{ly\_rumi}^n) \}$. For clarity, only the accuracy of the $F^n$ models is shown. \re{Red crosses indicate behavioural classes whose recall falls below the theoretical random-guessing reference of 0.17.} In general, the $F^n$ models demonstrate more stable and competitive performance than the corresponding $M^n$ models, particularly under data-scarce conditions.}
	\label{FIG:8}
\end{figure}

A unified model was developed which is capable of recognising all six target behaviours, particularly behaviours with extremely limited training samples. 
For clarity, consider the case where \textit{feeding} is treated as the data-scarce behaviour. The base classifier $B_{feed}$ is first trained using all available samples from the remaining five behaviours. Starting from $B_{feed}$, the final model $F_{feed}^n$ is then adapted using $n$ labelled \textit{feeding} samples together with $n$ randomly selected samples from each of the other five behaviours. As a baseline, $M_{feed}^n$ is trained directly using the same $n$ \textit{feeding} samples and all available samples from the remaining five behaviours. Both $F_{feed}^n$ and $M_{feed}^n$ are evaluated on the same test set containing all six behaviours. This procedure is repeated for each behaviour in turn.
\re{To evaluate class-wise performance under data-scarce conditions, a theoretical random-guessing reference was defined for recall. Since the task involves six behavioural classes, a uniform random classifier predicts each class with probability $\frac{1}{6}$. Therefore, the expected recall for any given class is approximately $\frac{1}{6} \approx 0.17$. Behavioural classes with recall below this reference value are marked with a red cross in Figure~\ref{FIG:8}.}

\begin{figure}[H]%[!thbp]
%	\centering
    \includegraphics[width=.98\textwidth]{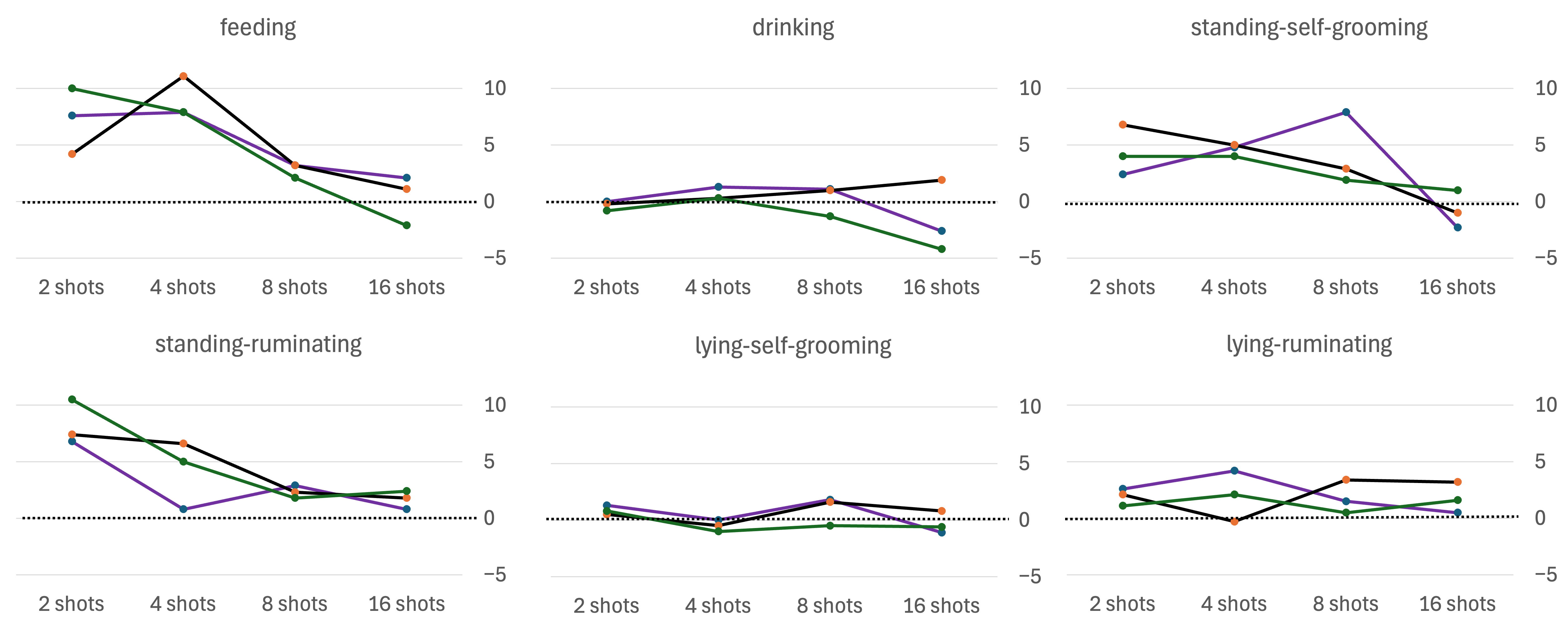}
  % \vspace{0.5mm}
    \caption{{Performance} %\textcolor{( 1. Please confirm whether an explanation of the colorful lines with dots needs to be added to the figure caption. 2. Please revise the hyphen to be minus sign (−) in figure.)}
 difference between the final model and the corresponding base model across three dataset splits. Each curve represents one dataset split produced using a different random seed, illustrating the sensitivity of the results to data partitioning. 
 The proposed base-to-novel transfer learning strategy yields the largest and most consistent improvements for \textit{feeding}, \textit{standing-ruminating} and \textit{lying-ruminating}, particularly under low-shot conditions, whereas \textit{drinking} and \textit{lying-self-grooming} exhibit little evidence of consistent improvement, with performance differences remaining close to zero across dataset splits.}
	\label{FIG:delta}
\end{figure}

% overall performance
As illustrated in Figures~\ref{FIG:8} and \ref{FIG:delta}, the proposed few-shot models $F^n$ (orange curves) achieve competitive or superior recognition performance compared with the corresponding one-stage baseline models $M^n$ (blue curves), particularly under low-shot settings. 
Although the performance differences are relatively small for behaviours such as \textit{drinking} and \textit{lying-self-grooming}, the proposed framework demonstrates substantially improved robustness as the number of training samples decreases. 
In several behavioural categories, the baseline models exhibit noticeable performance degradation when trained with only a few samples, whereas the proposed few-shot models maintain relatively stable recognition capability.

% robustness
This robustness is particularly evident in the \textit{standing-self-grooming} and \textit{standing-ruminating} tasks. 
As the number of training samples decreases, the performance of the proposed models declines gradually, indicating stronger tolerance to data scarcity. 
In contrast, the corresponding one-stage baseline models suffer from much sharper performance drops and may even fail to maintain effective recognition capability under extremely limited training conditions. 
Furthermore, many one-stage models $M^2$ become unstable when only two samples are available, and even $M_{ly\_groom}^4$ fails to achieve effective recognition performance. 
By comparison, the proposed few-shot models $F^2$ still preserve effective recognition capability across five behavioural categories, demonstrating stronger resistance to severe class imbalance and insufficient training data.

\begin{figure}[H]
%	\centering
    \includegraphics[width=.98\textwidth]{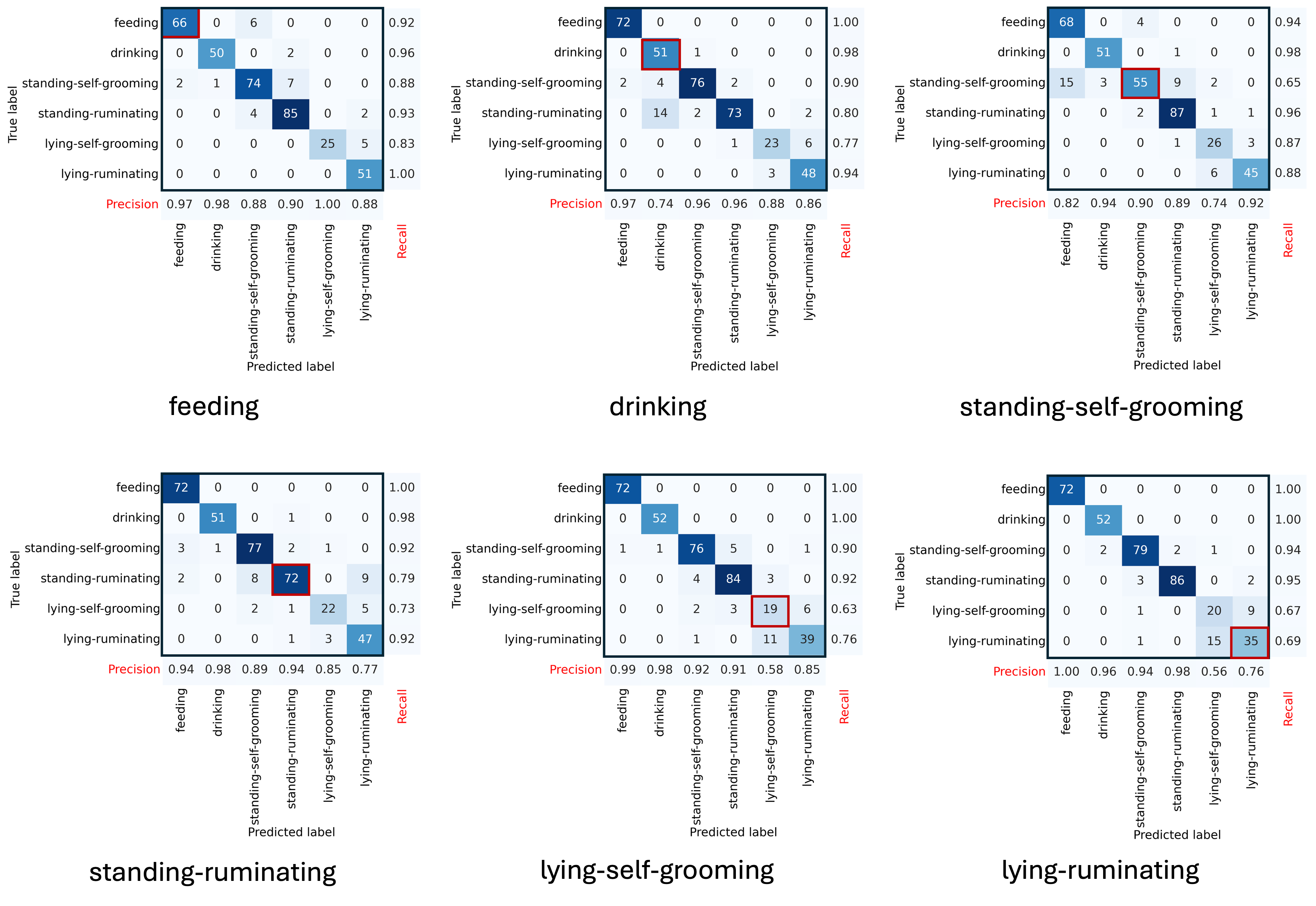}
    \caption{{Confusion} %\textcolor{( Please confirm whether the explanations of the red font and colors need to be added to the figure caption.)}
 matrices for the few-shot models $F_{feed}^{16},$ $F_{drink}^{16},$ $F_{std\_groom}^{16},$ $F_{std\_rumi}^{16},$ $F_{ly\_groom}^{16}$, $F_{ly\_rumi}^{16}$ on the first dataset split. The red boxes mark the results of data-scarce behaviours. With the exception of one underperforming variant, the $F^{16}$ models demonstrate strong capability in learning behaviours from limited data.}
	\label{FIG:7}
\end{figure}

\begin{figure}[H]%[!htbp]
	%\centering
    \includegraphics[width=.99\textwidth]{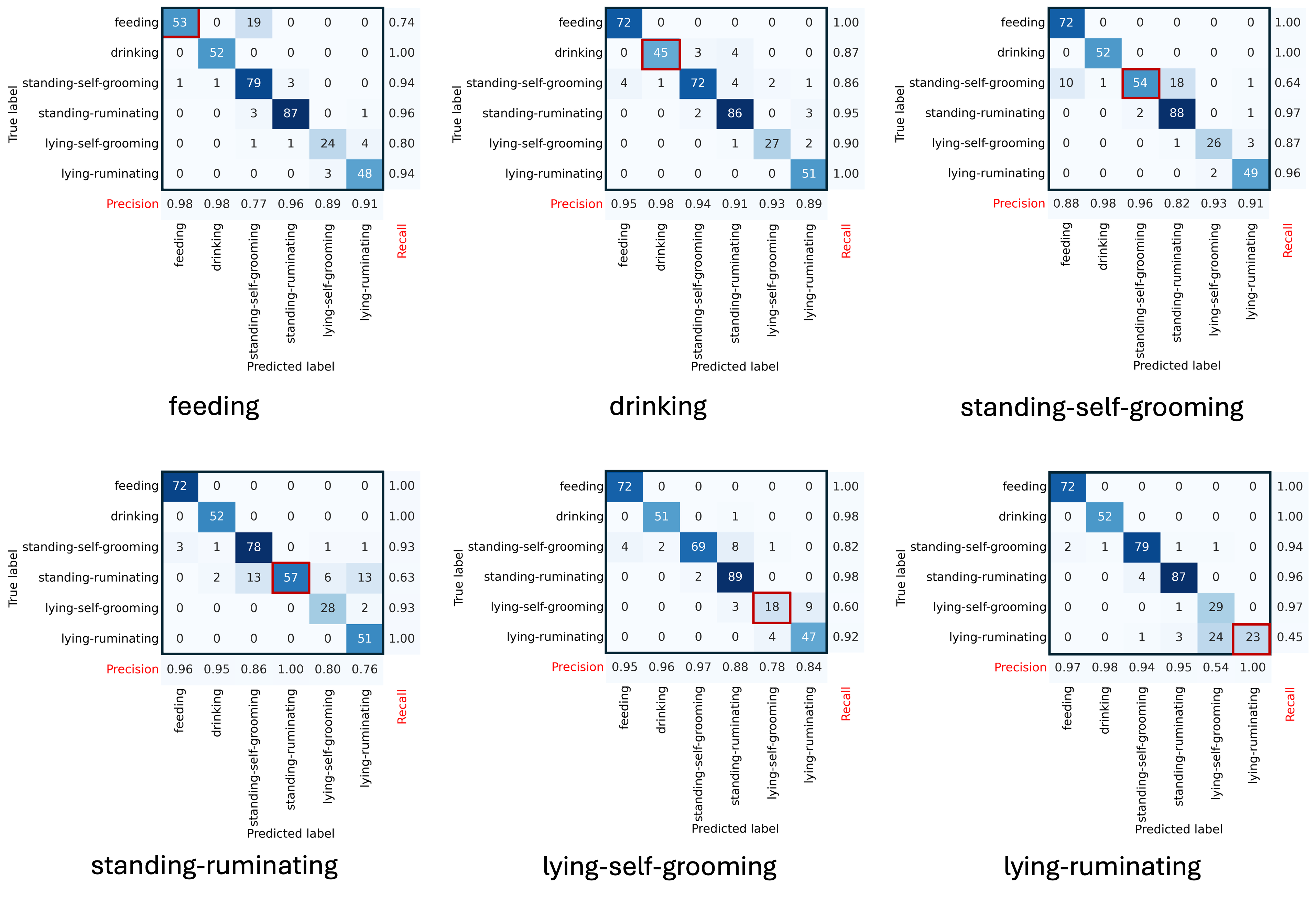}
    \caption{{Confusion} %\textcolor{(  Please confirm whether the explanations of the red font and colors need to be added to the figure caption.)}
 matrices of baselines $M_{feed}^{16},$ $M_{drink}^{16},$ $M_{std\_groom}^{16},$ $M_{std\_rumi}^{16},$ $M_{ly\_groom}^{16},$ $M_{ly\_rumi}^{16}$ on the first dataset split. The red boxes mark the results of data-scarce behaviours. The $M^{16}$ models tend to misclassify data-scare behaviours as other categories, resulting in low recall.}
	\label{FIG:9}
\end{figure}

% confusion matrix + recall
Recall is emphasised in this study because it better reflects the model's ability to identify under-represented novel behaviours. 
Comparison of the confusion matrices in {Figures} %\textcolor{( We merged Figure 7 and Figure 11 into Figures 7 and 11 there, please confirm.)}
 \ref{FIG:111} and \ref{FIG:7} further demonstrates that the proposed base-to-novel few-shot learning framework with replay not only preserves knowledge learned from the base classifier but also improves generalisation to novel behaviours. 
For example, the model $F_{feed}^{16}$ in Figure~\ref{FIG:7} maintains strong recognition performance on \textit{drinking}, \textit{lying-self-grooming} and \textit{lying-ruminating} while simultaneously achieving high recognition performance on the data-limited \textit{feeding} behaviour, with a recall of $0.92$ and precision of $0.97$. 
In contrast, the corresponding baseline model $M_{feed}^{16}$ in Figure~\ref{FIG:9} struggles to distinguish \textit{feeding} from \textit{standing-self-grooming}, resulting in a substantially lower recall of $0.74$.

A similar trend can be observed in other data-limited behaviours. 
For instance, $M_{std\_rumi}^{16}$ and $M_{ly\_rumi}^{16}$ achieve recalls of only $0.63$ and $0.45$, whereas the corresponding few-shot models $F_{std\_rumi}^{16}$ and $F_{ly\_rumi}^{16}$ improve the recalls to $0.79$ and $0.69$, respectively. 
Although the corresponding F1-scores (Table~\ref{tab:f1_score_fm}) do not always increase proportionally because of the trade-off between precision and recall, the improved recall indicates that the proposed framework is more effective at detecting rare and under-represented behaviours. 
These results suggest that the one-stage models tend to overfit to the dominant base behaviours under severe class imbalance, whereas the proposed replay-based few-shot framework achieves better balance between preserving base knowledge and adapting to novel behavioural categories.

The lack of observable performance differences in recognising \textit{drinking} and \textit{lying-self-grooming} behaviours can be explained by the intrinsic similarity between these behaviours and other dominant categories in the dataset. 
As shown in Figure \ref{FIG:7}, \textit{drinking} is occasionally confused with \textit{standing-ruminating}, since some \textit{standing-ruminating} samples are collected near the drinking trough and therefore share similar visual characteristics. 
Similarly, \textit{lying-self-grooming} is frequently misclassified as \textit{lying-ruminating} because the two behaviours exhibit highly similar body postures and spatial appearance. 
When more \textit{lying-ruminating} samples are seen, the model tends to develop a bias towards the dominant \textit{lying-ruminating} category, making the under-represented \textit{lying-self-grooming} behaviour more difficult to distinguish. 
In contrast, when the number of \textit{lying-ruminating} samples is reduced, the class imbalance between the two behaviours becomes less severe, which alleviates the dominance effect and leads to relatively more stable recognition performance.

\vspace{-12pt}

\begin{table}[H]
    \caption{{Comparison} %\textcolor{(  Please add an explanation for the use of bold in the table footer. If the bold is unnecessary, please remove it in table body. )}
 of per-class F1-score between few-shot models $F^{16}$ and baseline models $M^{16}$ on the first dataset split.
 \label{tab:f1_score_fm}}
    \begin{adjustwidth}{-\extralength}{0cm}
 %   \centering
    \small
    \begin{tabularx}{\fulllength}{@{}lCCCCCC@{}}
    \toprule
    \textbf{Behaviours} & \textbf{Feeding} & \textbf{Drinking} & \textbf{Standing-Self-Grooming} & \textbf{Standing-Ruminating} & \textbf{Lying-Self-Grooming} & \textbf{Lying-Ruminating} \\
    \midrule
    $F_{feed}^{16}$ & 0.94 & 0.97 & 0.88 & 0.92 & 0.91 & 0.94 \\ 
    $M_{feed}^{16}$ & 0.84 & 0.99 & 0.85 & 0.96 & 0.84 & 0.92 \\
    \midrule
    $F_{drink}^{16}$ & 0.99 & 0.84 & 0.93 & 0.87 & 0.82 & 0.90 \\ 
    $M_{drink}^{16}$ & 0.97 & 0.92 & 0.89 & 0.92 & 0.92 & 0.94 \\
    \midrule
    $F_{std\_groom}^{16}$ & 0.88 & 0.96 & 0.76 & 0.92 & 0.80 & 0.90 \\ 
    $M_{std\_groom}^{16}$ & 0.94 & 0.99 & 0.77 & 0.89 & 0.90 & 0.93 \\
    \midrule
    $F_{std\_rumi}^{16}$ & 0.97 & 0.98 & 0.90 & 0.86 & 0.79 & 0.84 \\ 
    $M_{std\_rumi}^{16}$ & 0.98 & 0.97 & 0.89 & 0.77 & 0.86 & 0.86 \\
    \midrule
    $F_{ly\_groom}^{16}$ & 0.99 & 0.99 & 0.91 & 0.92 & 0.60 & 0.80 \\ 
    $M_{ly\_groom}^{16}$ & 0.97 & 0.97 & 0.89 & 0.93 & 0.68 & 0.88 \\
    \midrule
    $F_{ly\_rumi}^{16}$ & 1.00 & 0.98 & 0.94 & 0.96 & 0.60 & 0.72 \\ 
    $M_{ly\_rumi}^{16}$ & 0.99 & 0.99 & 0.94 & 0.95 & 0.69 & 0.62 \\
    \bottomrule
    \end{tabularx}
    \end{adjustwidth}
\end{table}

%%%%%%%%%%%%%%%%%%%%%%%%%%%%%%%%%%%%%%%%%%
\section{Discussion}

\subsection{Main Findings and Contributions}
The results demonstrate the potential of multimodal vision--language learning for video-based dairy cow behaviour recognition. In contrast to conventional approaches that rely primarily on visual features and task-specific classifiers, Cattle-CLIP integrates temporal video representations with behaviour-relevant textual semantics. This enables the model to exploit semantic knowledge from multimodal pretraining while adapting it to the specific visual characteristics of cattle surveillance videos.
{These results can be contextualised against previous cattle behaviour recognition studies, although direct numerical comparisons should be interpreted with caution because of differences in datasets, behaviour definitions, and sensing modalities.
Guo et al. \citep{guo2019detection} reported an accuracy of 98.3\% for mounting behaviour using a dataset of 30 videos, whereas Ma et al. \citep{ma2022basic} achieved 95\% overall accuracy for the recognition of lying, standing and walking using an enhanced R(2+1)D-based model. Zhang et al. \citep{zhang2024new} further reported 98.80\% accuracy across five behaviours by combining video with neck-mounted IMU measurements. Collectively, these studies show that high recognition performance can be achieved for single behaviours, relatively coarse-grained behavioural categories, or systems incorporating additional sensor information.
By contrast, distinguishing posture-dependent behaviours that share similar visual characteristics presents a more fine-grained recognition problem, particularly when only video data are available. The comparison therefore highlights the importance of considering behavioural granularity and sensing modality, rather than absolute accuracy alone, when evaluating cattle behaviour recognition methods.}

{Beyond supervised recognition, the base-to-novel experiments further examine model adaptation when labelled data for particular behaviours are limited. 
The challenge of limited behavioural annotations has also been recognised in previous cattle studies. \mbox{Bloch et al. \citep{bloch2023development}}, for example, showed that transfer learning improved dairy cow behaviour classification from accelerometer data when only a limited amount of training data was available. Similarly, Mao et al. \citep{mao2024cross} demonstrated that transferring knowledge from sheep and horse activity data improved cattle behaviour recognition under limited labelled data. 
These studies indicate that knowledge transfer can reduce the dependence of cattle behaviour recognition models on extensive annotations. However, published evidence on few-shot learning for video-based cattle behaviour recognition remains limited, particularly for the recognition of under-represented behaviour categories.
The present base-to-novel experiments address this setting by examining whether knowledge acquired from well-represented behaviours can support the recognition of under-represented behaviours while retaining performance on previously learned classes.}

Together with the public release of CattleBehaviours6, these contributions provide both a methodological framework and a reusable benchmark for future cow behaviour recognition research.

\subsection{Behaviour-Specific Performance and Practical Implications}
{Analysis of the confusion matrices (Figure~\ref{FIG:111}) reveals disparities in class-wise performance.
While the model achieves high precision and recall for \textit{feeding} and \textit{drinking}, it encounters difficulty in distinguishing between \textit{lying-self-grooming} and \textit{lying-ruminating}. Similar patterns emerge in the few-shot experiments, where 29\% and 47\% of \textit{lying-ruminating} samples are misclassified as \textit{lying-self-grooming} by $F_{std\_rumi}^{16}$ (Figure~\ref{FIG:7}) and $M_{std\_rumi}^{16}$ \mbox{(Figure~\ref{FIG:9})}, respectively.
These differences are likely related to the visual and temporal characteristics of the behaviours. \textit{Feeding} and \textit{drinking} involve distinctive postures and interactions with environmental structures, making them comparatively easier to distinguish. Attention maps (Appendix~\ref{app:attention}) further indicate that Cattle-CLIP uses both animal-related features and behaviour-relevant environmental context. Because such contextual cues form part of the operational definitions of these behaviours, they provide meaningful information for recognition, although reliance on them may reduce robustness when recording environments change substantially.
In contrast, \textit{lying-self-grooming} and \textit{lying-ruminating} share similar body postures and differ mainly in subtle head and mouth movements. Because the current framework aggregates frame-level features using average pooling, temporal information is represented only implicitly. This may limit recognition of fine-grained motion patterns. This differs from previous video-based cattle behaviour studies focused on more distinct postural or locomotion states. For example, Ma et al. \citep{ma2022basic} reported 95\% overall accuracy for lying, standing and walking, behaviours characterised by clearer differences in posture and locomotion. The confusion observed in the present study therefore suggests that fine-grained within-posture recognition is more challenging than coarse posture classification and may benefit from more explicit temporal modelling.
From a practical perspective, hybrid systems combining wearable sensors and video have reported very high recognition performance, such as the 98.80\% accuracy reported by Zhang et al. \citep{zhang2024new}, but require additional sensing devices to be attached to animals. In contrast, video-based monitoring can provide continuous behavioural information without wearable hardware. The ability to adapt the model using relatively limited labelled data may also facilitate deployment in new operational settings where extensive annotation is impractical. In the longer term, reliable recognition of changes in behaviours such as \textit{feeding}, \textit{drinking} and \textit{rumination} could support automated monitoring of behavioural deviations associated with health or welfare changes. However, the present study evaluates behaviour recognition rather than health-event detection, and such applications will require further longitudinal validation.}

\subsection{Limitations}
{Several limitations should be considered when interpreting the present results. First, persistent individual identities were not available, and therefore the current clip-level evaluation does not demonstrate generalisation to completely unseen cows. Second, this study was conducted on Holstein--Friesian cows within a single farm environment, and performance under different breeds, housing systems, camera viewpoints and backgrounds remains to be established. Third, the current ethogram contains six mutually exclusive behaviours, where some cattle behaviours may occur concurrently and some categories are distinguished primarily by subtle temporal motion cues, while performance under highly data-constrained few-shot settings remains sensitive to the selected adaptation samples. Finally, to translate our approach from a research tool to industrial adoption, assessment and optimisation on edge devices will be essential.}

\subsection{Future Directions}
\textbf{Robustness and generalisation.}
{The most immediate priority is to establish whether Cattle-CLIP generalises beyond the current clip-level, single-farm evaluation. 
Building on our previous cattle re-identification methods developed and evaluated in the same farm environment \citep{yu2025holstein, andrew2021visual}, future work will evaluate the behaviour recognition performance of Cattle-CLIP on unseen cows.
We also aim to assess the generalisation performance across multiple farms with different production systems and breeds and need for few-shot adaptation using small labelled subsets from the target domains, as illustrated in this paper.
The main practical requirements are additional identity-labelled data covering diverse camera locations and sufficiently standardised annotation protocols across farms.}

\textbf{Expanded ethogram and temporal modelling.}
{In parallel, our aim is to realise our approach as a tool to underpin AI-driven veterinary research. For this, it is necessary to extend the behavioural scope of the dataset and improve recognition of fine-grained behaviours. 
As a first step, we are currently developing a comprehensive cattle ethogram with standardised operational definitions for each behaviour to support consistent annotation and future dataset expansion. The annotation scheme will be extended to represent concurrent and compound behaviours. For example, posture and activity could be annotated separately, allowing combinations such as \textit{lying + ruminating} or \textit{standing + self-grooming} to be represented explicitly rather than treated only as predefined compound classes. Nevertheless, the optimal modelling strategy is currently an open question---for example, should all compound classes be enumerated and learnt by the model in a conventional multiclass classification framework, or should the model predict concurrent aspects, e.g., posture and activity, separately.}  

{As the ethogram becomes larger and more behaviourally detailed, temporal modelling is expected to become increasingly important. The current Cattle-CLIP framework uses average pooling to aggregate frame-level representations. A subsequent experiment will therefore replace this component with temporal-attention or dedicated spatio-temporal fusion modules and compare them under otherwise consistent training and evaluation settings. Particular attention will be given to behaviour pairs such as \textit{lying-self-grooming} and \textit{lying-ruminating}, which share similar postures but differ in subtle head and mouth movements. Temporal modelling approaches that have proved effective in human action recognition could provide useful architectural starting points for this investigation \citep{wang2024multimodal,zhang2025efficient}.
A principal challenge is that many cattle behaviours are considerably more fine-grained than conventional human action-recognition categories. Their distinction may depend on subtle, localised and repetitive movements rather than large changes in body posture or locomotion. Consequently, extending the ethogram will require not only additional annotated data but also temporal representations capable of preserving these fine-scale motion patterns.}

%%%%%%%%%%%%%%%%%%%%%%%%%%%%%%%%%%%%%%%%%%
\section{Conclusions}

This work explored the integration of multimodal pretraining into livestock behaviour recognition through the proposed Cattle-CLIP framework. By incorporating lightweight temporal modelling and domain-aware adaptation strategies, this study demonstrates the feasibility of transferring vision–language representations to video-based cow behaviour analysis.
Beyond supervised evaluation, the base-to-novel experiments further demonstrate the potential of multimodal semantic supervision for behaviour recognition when labelled examples are limited.
In addition, CattleBehaviours6 is released as a curated video dataset with standardised behavioural definitions and benchmark partitions. Together, the proposed framework and dataset provide a basis for further research into scalable and data-efficient automated cow behaviour monitoring.

%%%%%%%%%%%%%%%%%%%%%%%%%%%%%%%%%%%%%%%%%%

\vspace{6pt}

\authorcontributions{Conceptualisation, N.W.C. and A.W.D.; methodology, H.L. and D.B.; software, Huimin Liu and A.X.M.; validation,  H.L.; investigation,  H.L.; data curation,  H.L. and Daria Baran; writing---original draft preparation,  H.L.; writing---review and editing,  H.L., J.G., N.W.C. and A.W.D.; supervision, N.W.C. and A.W.D.; project administration, A.W.D.; funding acquisition, N.W.C. and A.W.D. All authors have read and agreed to the published version of the manuscript.}

\funding{{Huimin} %\textcolor{(  Information regarding the funder and the funding number should be provided. Please check the accuracy of funding data and any other information carefully.)}
 Liu is supported by the China Scholarship Council (202208420042). Andrew Dowsey, Neill Campbell, Axel Montout and Jing Gao received support from Biotechnology and Biological Sciences Research Council (BBSRC) under grant ID BB/X017559/1, which also covered Open Access publication costs.}

\institutionalreview{\re{The study was carried out in accordance with the Declaration of Helsinki and was approved by the Animal Welfare and Ethical Review Body of University of Bristol (protocol code UIN-23-060, approved on 27 September 2023).}}

\informedconsent{\re{Not applicable.}}
% \informedconsent{Any research article describing a study involving humans should contain this statement. Please add ``Informed consent was obtained from all subjects involved in the study.'' OR ``Patient consent was waived due to REASON (please provide a detailed justification).'' OR ``Not applicable'' for studies not involving humans. You might also choose to exclude this statement if the study did not involve humans.

% Written informed consent for publication must be obtained from participating patients who can be identified (including by the patients themselves). Please state ``Written informed consent has been obtained from the patient(s) to publish this paper.” if applicable.}

\dataavailability{\re{The original data presented in the study are openly available in the CattleBehaviours6 {GitHub (version 1.0), accessed on 23 September 2026, } %\textcolor{( Please state which version of the software was used.)}
 repository {at} %\textcolor{red}{}
~\href{https://github.com/SolanLiu25/Dataset_CattleBehaviours6}{https://github.com/SolanLiu25/Dataset\_CattleBehaviours6} upon acceptance of the manuscript.}}

\acknowledgments{{The} %\textcolor{( Please ensure that all individuals included in this section have consented to the acknowledgement.)}
 authors would like to acknowledge the John Oldacre Centre for Dairy Welfare and Sustainability Research at the Bristol Veterinary School for its support of this work. For the purpose of open access, the authors have applied a Creative Commons Attribution (CC BY) licence to any Author Accepted Manuscript arising from this submission. \re{None of the figures or the graphical abstract in our manuscript were generated using artificial intelligence (AI) tools.}}

\conflictsofinterest{The authors declare no conflicts of interest.}

\appendixtitles{yes}
\appendixstart
\appendix
\section[\appendixname~\thesection]{Attention Map Visualisations}\label{app:attention}
To examine the visual cues used by Cattle-CLIP for behaviour recognition, attention maps were generated for representative \textit{feeding} and \textit{drinking} samples.
These visualisations illustrate the regions contributing to the model predictions and provide additional insight into the role of animal-related features and environmental context.

\begin{figure}[H]
 %   \centering
    \includegraphics[width=.99\textwidth]{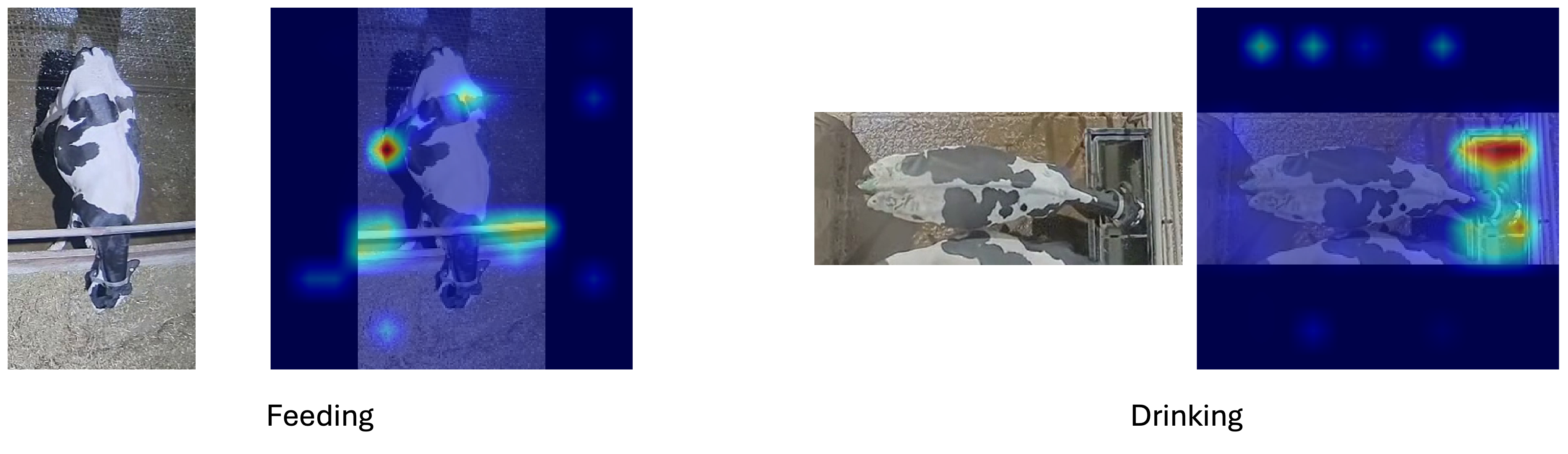}
    \vspace{-6pt}
    \caption{\re{{Attention} %\textcolor{( Please confirm whether an explanation of the colors needs to be added to the figure caption.)}
 map visualisations for representative feeding and drinking samples.
    The model attends to both animal-related regions and behaviour-relevant contextual
    areas surrounding the \textit{feeding} and \textit{drinking} locations.}}
    \label{FIG:attention_map}
\end{figure}

\section[\appendixname~\thesection]{Camera-Specific Evaluation}\label{app:camera}
\begin{figure}[H]
 %   \centering
    \includegraphics[width=.99\textwidth]{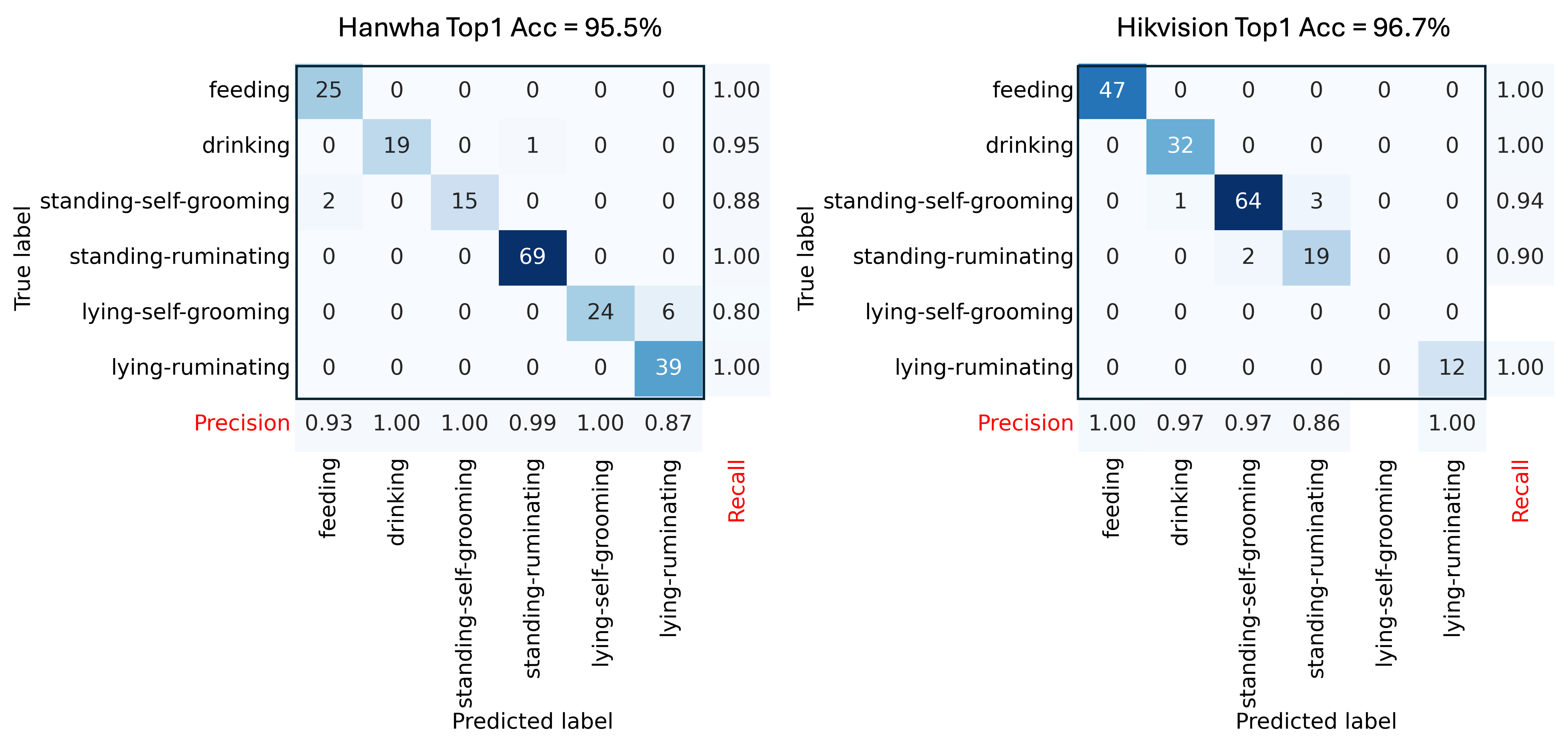}
    \caption{\re{{Confusion} %\textcolor{( Please confirm whether the explanations of the red font and colors need to be added to the figure caption.)}
 matrices of Cattle-CLIP evaluated separately on samples captured by the two camera types. Comparable overall performance was observed across the two camera groups.}}
    \label{FIG:camera_test}
\end{figure}

To investigate whether model performance differed between the two camera types used for data collection, an additional camera-specific evaluation was performed \mbox{(Figure~\ref{FIG:camera_test}).} The two camera groups achieved comparable overall performance, with accuracies of approximately 95.5\% and 96.7\%, respectively, suggesting no substantial camera-specific inconsistency under the conditions evaluated in this study. 
However, because camera type was associated with fixed recording locations, any observed differences may reflect not only camera hardware but also variations in viewpoint, background, illumination and behaviour distribution.
 \vspace{-3pt}

\section[\appendixname~\thesection]{Recording Schedule}\label{app:schedule}
Data collection was conducted over 20 non-consecutive recording days between
October 2024 and April 2025. Table~\ref{tab:recording_dates} summarises the
recording dates associated with each camera ID used in this study.

\begin{table}[H]
  %  \centering
    \caption{\re{{Recording} %\textcolor{( We added table lines to make the table content to be shown more clear in the table, please confirm. )}
 dates associated with the camera sources used for data collection.
    Recordings were collected over 20 non-consecutive days between October 2024
    and April 2025.}}
    \label{tab:recording_dates}
    \begin{tabularx}{\textwidth}{ll}
    \toprule  
    \textbf{Camera ID} & \textbf{Recording Dates} \\
    \midrule 

    1 &
    \makecell[l]{
    28 {December}%\textcolor{( We completed the month of date, please confirm the revisions in whole table. )}
 2024; 20 February 2025; 21 February 2025
    } \\\midrule

    2 &
    \makecell[l]{
    27 October 2024; {1} %\textcolor{( We removed 0 of day, please confirm the revisions in whole table. )}
 November 2024; 5 November 2024; 11 November 2024; \\
    12 November 2024; 13 November 2024; 21 February 2025
    } \\%\midrule

 \bottomrule 
    \end{tabularx}
\end{table}

\begin{table}[H]\ContinuedFloat
%\tablesize{\small}
\caption{\textit{Cont.}}
%\begin{adjustwidth}{-\extralength}{0cm}
%\centering
 \begin{tabularx}{\textwidth}{ll}
    \toprule  
    \textbf{Camera ID} & \textbf{Recording Dates} \\
    \midrule 
    3 &
    \makecell[l]{
    26 December 2024; 27 December 2024; 28 December 2024; 5 January 2025;\\ 21 February 2025
    } \\
    
\midrule
    4 &
    \makecell[l]{
    16 January 2025; 21 February 2025; 31 March 2025; 1 April 2025; \\
    3 April 2025; 4 April 2025; 20 April 2025
    } \\\midrule

    5 &
    \makecell[l]{
    28 December 2024; 5 January 2025; 21 February 2025; 31 March 2025; \\
    1 April 2025; 3 April 2025; 4 April 2025; 20 April 2025
    } \\\midrule

    6 &
    \makecell[l]{
    28 December 2024; 8 January 2025; 20 February 2025; 21 February 2025;\\
    } \\\midrule
    
    7 &
    \makecell[l]{
    10 November 2024; 20 February 2025; 21 February 2025
    } \\

    \bottomrule 
    \end{tabularx}
\end{table}

\begin{adjustwidth}{-\extralength}{0cm}

\reftitle{References}

\PublishersNote{}
%\isPreprints{}{% This command is only used for ``preprints''.
\end{adjustwidth}

%%%%%%%%%%%%%%%%%%%%%%%%%%%%%%%%%%%%%%%%%%
% \newpage

\end{document}